\pdfoutput=1

\documentclass[11pt]{article}

\usepackage[]{acl}

\usepackage{times}
\usepackage{latexsym}

\usepackage[T1]{fontenc}

\usepackage[utf8]{inputenc}

\usepackage{microtype}

\usepackage{supertabular}

\usepackage[normalem]{ulem}

\usepackage{amsmath}
\usepackage{graphicx}
\usepackage{booktabs}
\usepackage{multirow}
\usepackage{caption}
\usepackage{subcaption}
\usepackage{xcolor}
\usepackage{enumitem,kantlipsum}

\title{Discharge Summary Hospital Course Summarisation of In Patient Electronic Health Record Text with Clinical Concept Guided Deep Pre-Trained Transformer Models}

\author{Thomas Searle$^1$, Zina Ibrahim$^1$, James Teo$^2$, Richard JB Dobson$^{1,3}$ \\
  $^1$Department of Biostatistics and Health Informatics, Institute of Psychiatry, \\ Psychology and Neuroscience, King’s College London, London, U.K.\\
  $^2$King's College Hospital NHS Foundation Trust, London, UK \\
  $^3$Institute of Health Informatics, University College London,\\
  London, London, U.K.}

\begin{document}
\maketitle
\begin{abstract}

Brief Hospital Course (BHC) summaries are succinct summaries of an entire hospital encounter, embedded within \textit{discharge summaries}, written by senior clinicians responsible for the overall care of a patient. Methods to automatically produce summaries from inpatient documentation would be invaluable in reducing clinician manual burden of summarising documents under high time-pressure to admit and discharge patients. Automatically producing these summaries from the inpatient course, is a complex, multi-document summarisation task, as source notes are written from various perspectives (e.g. nursing, doctor, radiology), during the course of the hospitalisation. We demonstrate a range of methods for BHC summarisation demonstrating the performance of deep learning summarisation models across extractive and abstractive summarisation scenarios. We also test a novel ensemble extractive and abstractive summarisation model that incorporates a medical concept ontology (SNOMED) as a clinical guidance signal and shows superior performance in 2 real-world clinical data sets.
\end{abstract}

\section{Introduction}

A patient's clinical journey is documented in rich free-text narratives stored in time-ordered linked documents in Electronic Health Records (EHRs). Narratives include commentary from multiple care teams, specialisms and perspectives with varying scope, detail, structure and time-span covered. Content broadly presents the patient experience, symptoms, findings and diagnosis alongside resulting procedures and interventions. Clinical and social histories and future prognoses are often referenced to provide further context and any potentially follow up actions to occur in some defined time period. Single notes also often mention or refer to previous notes. An encounter such as a simple routine outpatient procedure could generate only a few sentences, whereas a complex admission may result in hundreds of distinct documents. When a patient is discharged from an inpatient encounter, the discharging clinician \textit{summarises} the entirety of the visit often within a section of the \textit{Discharge Summary} note known as the \textit{Brief Hospital Course} (BHC) section. For short, i.e. day case admissions BHC sections are likely to be short and potentially not clearly defined. For longer, multi-day, complex admissions where a patient is being discharged to a primary, community or even tertiary care service this section is more likely to be present as its vital for continuity of care\cite{Silver2022-vi}. However, with most free-text clinical narrative, there can be large variability with how this data is presented\cite{Sorita2021-gj}. Overall, it is generally accepted that an effective discharge summary should document the clinical events of an admission\cite{Ming2019-ix}.

\begin{figure}
    \centering
    \includegraphics[scale=0.47]{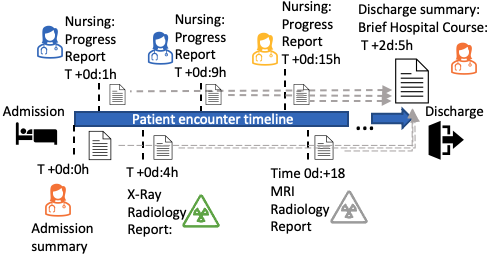}
    \caption{An example patient admission timeline where a patient is admitted with an admission summary note, nursing progress notes, radiology reports and a discharge summary note. Each is written by potentially different authors (colour coding), as the admission progresses. Each note potentially informs the BHC section within the discharge summary.}
    \label{fig:patient-encounter-timeline}
\end{figure}

Manually generating this summary is laborious, time-consuming and potentially error prone\cite{ODonnell2009-ye}. Fig. \ref{fig:patient-encounter-timeline} shows a fictitious, multi-day inpatient encounter. This single admission produces 6 distinct documents from a range of perspectives (Nursing, Doctors, Radiology) in the first 18 hours. The first 2 \textit{Nursing - Progress Notes} are by the same author, the differing radiology scans (X-ray vs. MRI) have different authors and the discharge summary is the same author that wrote the initial admission. Discharge occurs $\sim$2 days after admission with more notes taken than those shown. Each document can inform the BHC section. However, not all notes are treated equally, notes are categorised into care provider categories, and further by admission, progress, discharge amongst other types. Due to the volume of text and the time-constraints for doctors to produce these summaries, it is improbable that a clinician author reads the entirety of the record and certainly not thoroughly.

In computational linguistics, this problem can be framed as a challenging multi-document summarisation task, with the model required to adapt to varying numbers of documents (simple vs complex cases), large time variances between notes, differences between note types, varying source document authors aims and focus areas. 

A recent detailed analysis of BHC sections \cite{Adams2021-hp}, found BHC summaries to: 1) be information dense, 2) switch quickly between extractive and abstractive summarisation styles, beginning with top-level extractive summaries of an admission followed by \textit{problem orientated} abstractive summary of the admission, 3) be only a \textit{silver-standard} and can lack important information.

To the authors' knowledge, this is the first work to offer a range of summarisation models for BHC summarisation trained and tested on multiple real-world sources of clinical text. The contributions of this work are:
\begin{itemize}[leftmargin=*]
    \item A baseline evaluation of existing pre-trained Transformer models for abstractive summarisation fine-tuned on the BHC summarisation task.
    \item An evaluation of extractive top-k sentence extractive summarisation models. Using unsupervised and supervised methods to analyse the \textit{extractiveness} of the opening BHC sentences.
    \item An adapted abstractive summarisation model (BART)\cite{Lewis2020-pw} to include a clinical ontology aware guidance signal of relevant terms to produce \textit{problem-list} orientated abstractive summaries.
    \item An evaluation of an ensemble model for extractive and abstractive summarisation combining the extractive and abstractive models.
\end{itemize}

\section{Background}
\subsection{Automatic Summarisation}
Automatic summarisation of text aims to provide a concise, fluent representation of the source material, retaining `important' information whilst ignoring redundant or irrelevant information. Formally, with single document summarisation, a set of documents $T = \{t_1, t_2, \ldots t_n\}$ we aim to find some function $f(T) = T'$ where $T' = \{t_1', t_2', \ldots t_n'\}$ the set of texts that maximise some parameters of an effective summary. These parameters can include: \textit{maximum length} that could vary according to use case, \textit{correctness} if the generated summaries are factually inline with source texts, \textit{completeness}, if the generated summary captures all important information from source texts, and \textit{fluency}, a often subjective measure of the writing quality of the generated summary\cite{Laban2020-fo}. In multi-document summarisation we have multiple texts for each sample $T = \{t_{1_{1\cdots i}}, t_{2_{1\cdots j}}, \ldots t_{n_{1\cdots k}}\}$. With BHC summarisation each $t_i$ has one or more documents.

\subsection{Extractive \& Abstractive Summarisation}
Research interest in automatic summarisation has a long history with empirical data-driven methods divisible into two main groups\cite{Orasan2019-dm}. 
\begin{enumerate}
    \item Extractive summarisation is the selection and combination of important words, phrases, or sentences i.e. some syntactic unit, of source texts to form the summary text. Consider some document text $t$ of syntactic units $S = \{s_1, s_2, \cdots s_3\}$, $f(t) = t'$ where $t' = S'$ and $S' \subset S$.
    
    Some extractive summarisation methods can be considered Information Extraction (IE)\cite{White2001-oz} methods that identify important information and simply use $s_j$ where the information is found, or possibly surrounding syntactic units $s_{j-1}$ and $s_{j+1}$. Information is extracted until desired summary length is reached or there is no more information to extract. Further extractive approaches search / rank a document's $S$ according to some \textit{importance} metric and select the top-n many sentences for the desired summary length\cite{Zhong2019-ax}.
    \item Abstractive summarisation methods do not enforce generated summaries to be directly drawn from source texts. Instead, abstractive methods allow $f$ to generate any syntactic unit, i.e. $S' \not\subset S_i$. This means a `generation' step is used once a latent \textit{importance} model of source texts $T$ is found. Models are often equipped with a suitable vocabulary $V$ and are tasked with generating fluent, informative summary text, whist being guided by the latent \textit{importance} model.
\end{enumerate}

Prior work has combined extractive and abstractive approaches, allowing $f$ to balance abstractive and extractive summarisation, most notably the pointer-generator model\cite{See2017-fx}. 

Recently, large pre-trained Transformer\cite{Vaswani2017-db} models have been shown to perform well across a range of tasks such as machine translation, question answering and abstractive summarisation\cite{Raffel2020-hb}. The Transformer architecture supports learning of deep latent representations of input data by layering \textit{encoder} and \textit{decoder} blocks, the model learns deep contextual representations of input, and how to decode these representations for a range of sequence-to-sequence tasks. 

\subsection{Clinical Text Summarisation}
Clinical narratives are estimated to comprise 80\% of EHR data\cite{Murdoch2013-wx}. However, the development and application of text summarisation methods is progressing slowly\cite{Mishra2014-np} when compared with areas such as disease prediction\cite{Wynants2020-ja}, mortality prediction\cite{Johnson2017-pr}, and clinical information extraction\cite{Kreimeyer2017-pm}. Contributing factors include: 1) the difficulty in collecting reference summaries\cite{Adams2021-hp}, \textit{Gold standard} reference summary collection is difficult as the language is complex and highly specialised, 2) produced summaries present a \textit{high stakes} AI scenario that has potential to cause negative downstream effects\cite{Sambasivan2021-mq} if the model makes errors, 3) assessing summarisation model performance using automated metrics such as ROUGE\cite{Lin2004-my} is difficult, as high scoring models can still perform poorly when assessed by human evaluators\cite{Sai2020-sv}. 

Prior work has initially focused on extractive approaches\cite{Moen2016-js}. Approaches focused on modelling semantic similarity, and methods to optimally pick representative sentences, i.e. $s_i$ units, from latent topics discovered during model fitting. Recent work, has focused on single document summarisation of radiology reports\cite{Zhang2018-im,Kondadadi2021-vi,Dai2021-ap}. Radiology reports generally consist of three sections, the \textit{background} of patient, the \textit{findings} - the visible phenomena within the scan and finally the \textit{impression} - an often abstractive summary of the background and findings used during the clinical followup. The \textit{impression} sections are treated as the target reference summaries for model development. Radiology report summarisation is similar to a single document open-domain task, where modelling sentence salience and sentence compression are the primary aims. 

\subsection{Consistency of Discharge Summary Content}
There is currently no standard for discharge summary format or content although a majority of surveyed clinicians agree there should be a standard\cite{Sorita2021-gj}. There is ongoing work to improve the consistency of education and training in effective discharge summary writing\cite{Ming2019-ix,Stopford2015-tt} and in some fields such as surgical pathology, synoptic reporting is an agreed upon standard form for reporting clinical events\cite{Renshaw2018-gw}. Unfortunately, there is no cross specialty, consistent method of writing a BHC, or even a consistent section header to define the BHC section of a discharge summary. We presume this is due to the variability in clinical encounters that are documented, where it would be very difficult to define rigid structure to cover all eventualities without being overly onerous.  

\subsection{BHC Section Analysis}
Prior work\cite{Adams2021-hp} has shown BHC sections are: 1) dense with clinical terms, 2) can vary widely in complexity and quality, 3) quickly switch between extractive and abstractive styles. These make the BHC summarisation task a difficult task. In this work we attempt to find an effective method that can be consistently used across these varied datasets and that takes advantage of the density of clinical terms. It is outside the scope of this work to address the issue of the variability in discharge summary and BHC sections themselves.

\section{Datasets \& Methods}

\subsection{Datasets}\label{sec:datasets}
We extensively pre-process and clean the admission's discharge summaries to extract only the BHC section. We discard the rest of the discharge summary so as to not bias the source texts.

Our datasets are:
\begin{itemize}
    \item MIMIC-III: \cite{Johnson2016-mq} A large, US based ICU dataset collected between 2001-2012 containing 47,591 unique patient admissions. We extract BHC sections from discharge summaries with regular expressions and clean all other notes of headers / footers resulting in 1,441,109 unique documents. 
    \item KCH: clinical records for inpatients diagnosed with cerebral infarction (ICD10 code:I63.*) from the King’s College Hospital (KCH) NHS Foundation Trust, London, UK, EHR. We extract data via the Trust deployed CogStack\cite{Jackson2018-km} system, an Elasticsearch based ingestion and harmonization pipeline for EHR data. We extract BHC sections with regular expressions and clean source notes of common headers / footers resulting in 34,179 unique documents.
\end{itemize}

Table \ref{tab:desc_stats} shows that the average case includes many documents, over a multi-day stay. The MIMIC-III dataset of USA based ICU admissions, are skewed towards complex multi-day stays generating many small EHR notes. The KCH dataset are UK-derived clinical records containing only patients diagnosed with cerebral infarction requiring inpatient rehabilitation for associated disability and therefore covers substantially longer time periods. 

Concatenating entire patient episode free-text narratives can create very long sequences of text. For encounters that are over 1000 sentences we pick the top and bottom 500 sentences, based on the intuition that patient notes often begin with an important admission note describing the patient history, initial diagnosis and finish with the most recent summary of the patient state.  Our source-code for cleaning and preparing the data, and the following model code is made available to the research community\footnote{https://github.com/tomolopolis/BHC-Summarisation}.

\setlength\tabcolsep{3pt}
\begin{table}[]
    \centering
    \begin{tabular}{cccccc}
        \toprule
         \textbf{Dataset} &  \textbf{\# Adm} &  \textbf{\shortstack{Adm\\Length}} &  \textbf{\# Docs} & \textbf{\shortstack{Src\\Seq}} & \textbf{\shortstack{BHC\\Seq}} \\\midrule
            \textbf{M-III} &  47,591  & 7 &  26 & 206 & 731 \\
            \textbf{KCH} &    1,586   & 49  & 21 & 441 & 274 \\\bottomrule
    \end{tabular}
    \caption{Descriptive statistics for our MIMIC-III (M-III) and KCH clinical text data. From left to right, the number of admissions, the average admission length in days, the average number of notes per admission, the average sequence length of a document excl. the discharge summary, and the average sequence length of the BHC section within the the discharge summary.}
    \label{tab:desc_stats}
\end{table}

\subsection{Assessing Model Performance}\label{sec:model_perf}
We assess performance of our models using ROUGE\cite{Lin2004-my}. The ROUGE authors describe ROUGE-recall to measure the generated summaries `coverage' of the reference summary or how much of the reference summary is included with the generated summary. ROUGE-precision measures relevancy, or how much of the generated summary is relevant to reference summary. An ideal summary will balance both coverage and relevancy, which can be expressed as the ROUGE-F1 score. A higher ROUGE score correlates with higher human levels of satisfaction with a generated summary but there are still notable issues with the score interpretation\cite{Schluter2017-ov} . For context, in open-domain summarisation tasks ROUGE often still used and reported. Current state-of-the-art performance is 37-41 points \cite{Lewis2020-pw}.

\subsection{Extractive Baseline BHC Approaches}
Our initial experiments test a recent finding that BHC sections are often extractive summaries initially before moving to more abstractive summaries as the BHC section progresses\citet{Adams2021-hp}. We compare a range of unsupervised and supervised extractive summarisation models to predict the initial sentences of the BHC sections.

All methods first concatenate each document text in chronological order, split into sentences via Spacy\footnote{https://spacy.io/}, then embed sentences by averaging GloVe\cite{Pennington2014-fn} or directly using S-BERT\cite{Reimers2019-ip} embeddings, finally feeding these to a ranking model, an unsupervised TextRank\cite{Mihalcea2004-kw} or supervised Bi-LSTM\cite{Hochreiter1997-uw} model. We train multiple models to select top 1 to 15 ranked sentences. Our final baseline model, the Oracle model, uses the reference summary to rank source sentences via Gestalt Pattern matching \cite{Black2004-jl} computing the ratio of matching `tokens' (i.e. white-space separated words), for each reference summary sentence and source sentence pair. The top \textit{k} ranked source sentences are used in the oracle. The Oracle model provides an estimate of the performance ceiling of sentence based extractive summarisation for both datasets. 

\subsection{Pre-Trained Transformer Based Models}
We consider end-to-end abstractive summarisation models as further baselines. Large pre-trained Transformer\cite{Vaswani2017-db} models have been successful across a variety of tasks including textual summarisation. Models such as BERT\cite{Devlin2019-ny}, T5\cite{Raffel2020-hb} and BART\cite{Lewis2020-pw} have demonstrated state-of-the-art performance across classification, summarisation, translation, language comprehension and question answering with for the most part a single model architecture. Transformer models for sequence-to-sequence (seq2seq) tasks such as machine translation and summarisation consist of layers of Transformer blocks configured either as \textit{encoders} or \textit{decoders}. Models such as T5 and BART are end-to-end trained for a range of tasks, whereas BERT in its original configuration consisted of encoder only Transformer blocks. Further work has showed BERT models can be repurposed in encoder-decoder configurations for summarisation\cite{Rothe2020-gq}.

Once trained on large, open-domain datasets these models can be re-used on further specialised domains, transferring base knowledge to a narrower domain and problem\cite{Rogers2020-nl}. Transfer learning has recently been shown to be effective for biomedical use cases\cite{Peng2019-cw}. However, to our knowledge BHC summarisation has not been considered to date, and our baseline experiments initially establish if large pre-trained models can be fine-tuned to produce high quality BHC sections from source notes directly. 

All abstractive models have been pre-trained on large corpora of open-domain text prior to fine-tuning with clinical text. We use existing pre-trained model parameter and configurations from the publicly available huggingface model hub\footnote{https://huggingface.co/models}. Then, Fine-tuning is performed using 3 Nvidia Titan X GPUs (M-III experiments) and 8 Nvidia DGX V100 GPUs (KCH experiments). We use different compute for each dataset due to the difference in availability of access, the Nvidia DGX machine is shared, and restricted infrastructure co-located on the KCH site whereas the other hardware is openly accessible on the university network. We split datasets 80/10/10 for training, validation and test. We fine-tune for 20 epochs assessing validation set performance after each epoch. Initial experiments determined learning rate schedule and optimiser parameters and a suitable number of epochs for convergence. We report our results in Section \ref{sec:results} on the test set only. 

As discussed in Section \ref{sec:datasets} clinical notes and BHC sections are highly variable in length and complexity. One limitation of recent models are the limited source and target text sequence lengths that can be produced due to the self-attention mechanism requiring all input representations to attend to all others. For example, BERT scales quadratically limiting the max input sequence length to 512. BHC summarisation is difficult as both input source notes are (far) greater than this maximum, as shown in Table \ref{tab:desc_stats}. Recent models such as the Reformer\cite{Kitaev2020-ax}, and LongFormer\cite{Beltagy2020-gt} use various optimisations for the attention calculations to enable longer sequences to be encoded / decoded.

Abstractive summarisation models use source text saliency to focus the summary on only the important parts of the source text. Models must also learn how to faithfully produce source texts alongside ensuring the correct content. Prior work has shown models can be prone to \textit{hallucinations}, producing text that is not representative of the source text\cite{Zhao2020-vd}. This is problematic for high risk settings such as healthcare but to our knowledge this problem has only been studied for radiology report summarisation\cite{Zhang2020-px}.

\subsection{Clinical Concept Guided Summarisation}
Guiding summarisation models using a variety of guidance stimulus forcing the model to focus on specific inputs has recently been shown to be beneficial for open-domain summarisation\cite{Dou2021-lt}. 

We guide our abstractive summariser to focus on summarising the clinical problems and associated interventions of each admission, as is often the method used by clinicians when writing the BHC section\cite{Adams2021-hp}. We perform named entity recognition (NER) and entity linking to extract and link SNOMED-CT\cite{Stearns2001-yt} terms, a standardised clinical terminology via a pre-trained MedCAT\cite{Kraljevic2021-ln} model that has been unsupervised trained on both MIMIC-III and KCH datasets for SNOMED-CT \textit{problems} i.e. clinical findings, disease, disorders, and \textit{interventions} i.e. procedures and drugs. Specific top level SNOMED-CT terms provided in Appendix Tables \ref{tab:prob_terms} and \ref{tab:interven_terms}. We use \textit{MetaCAT} models configured within MedCAT to contextualise extracted terms. Therefore, all extracted terms are patient-relevant (i.e. not familial history mentions), positive (i.e. not negated), and are classified as a diagnosis (i.e. not mentions of department name or clinical specialisms e.g. ``patient attended the stroke clinic'' would not annotate stroke as a diagnosis). 

Appendix Table \ref{tab:medcat_desc_stats} provides full descriptive statistics of extracted terms across source and BHC notes. An interesting measure `term density' the average number of all word tokens per each extracted concept. This follows analysis in prior work that showed differences in the density of clinically relevant information between note types\cite{Adams2021-hp}. For example, if a sentence were to contain 20 words describing a patient diagnosis, of which our MedCAT model extracts 5 clinical terms, this would provide a term density of 4, as there are 4 word tokens for each clinical term. If of the 5 clinical terms there are only 2 unique clinical terms this provides a unique term density of 10. We observe that for MIMIC-III and KCH datasets the Notes and BHC sections have similar SNOMED-CT term density (56 vs 52) and word token densities (26 vs 29), but  when considering unique terms the BHC sections have almost double the density of unique clinical terms (63 Notes vs 118 BHC) for M-III whereas for the KCH notes it is circa equivalent (at 32 Notes vs 35 BHC), indicating in the M-III dataset BHC sections quickly change from one clinical topic to another when compared to source notes. Redundancy within these datasets have been described in prior work\cite{Searle2021-hk}.

\begin{figure}
    \centering
    \includegraphics[scale=0.22]{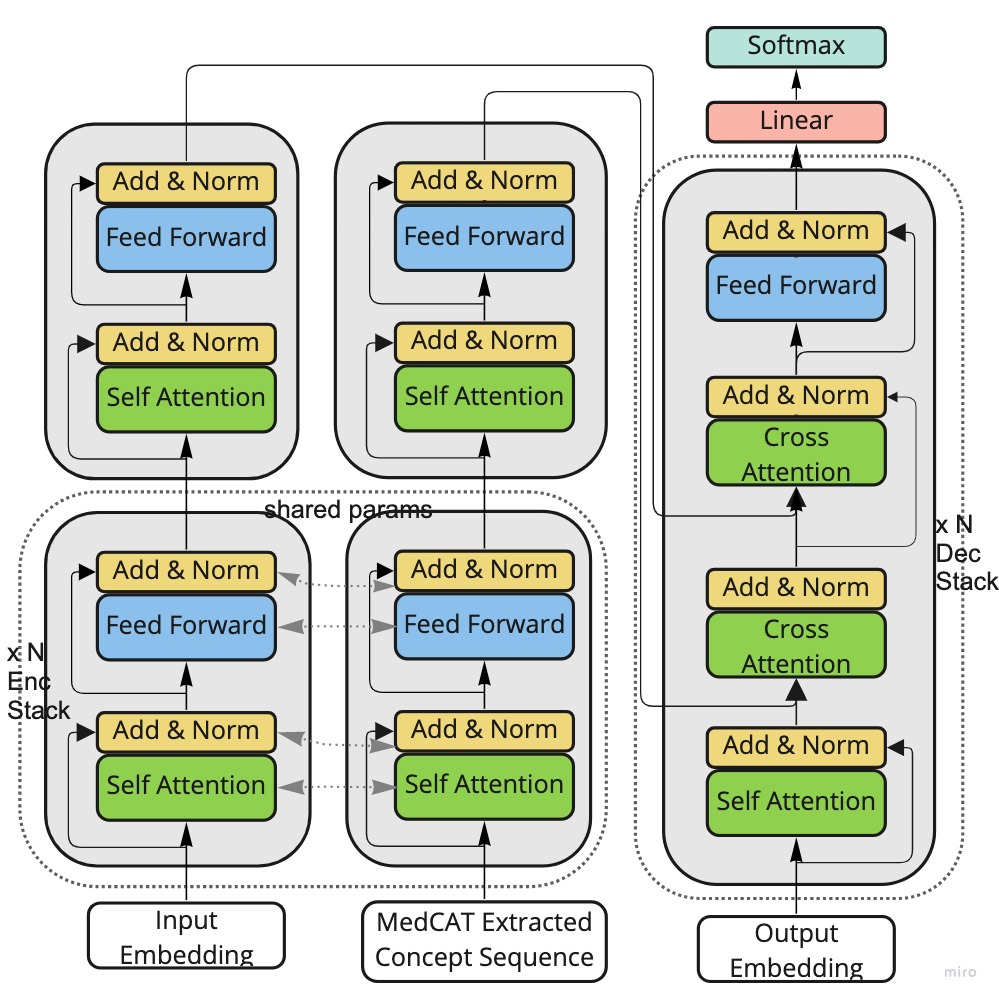}
    \caption{The Encoder-Decoder Architecture using Clinical relevant guidance signal during the encoding, decoding process. }
    \label{fig:problem-list-guided-arch}
\end{figure}

We use the huggingface\footnote{https://huggingface.co/} BART\cite{Lewis2020-pw} architecture pretrained on open-domain texts and additionally pretrained on a summarisation corpus PubMed\cite{Gupta2021-au}. We choose this architecture as it is specifically tuned for natural language generation (NLG) including summarisation. We follow the architecture modifications outlined in recent work\cite{Dou2021-lt}. This includes using dual Transformer based encoders, one for the raw text input and another for the MedCAT extracted guidance input. Importantly, the guidance input is aligned to the text input by padding guidance input, so the dual encoders receive the text and and MedCAT extracted concept term at the same sequence step. The model fails to converge without this alignment. These pre-trained parameters are shared for the first 3 encoder Transformer blocks reducing number of model parameters. The rest of the encoder Transformer blocks only see either the original text input or the associated guidance signal. The decoder Transformer blocks is implemented to use an extra cross-attention layer that uses the encoded guidance aware signal to the regular cross-attention layer from the text input encoder representation. Figure \ref{fig:problem-list-guided-arch} shows our clinical concept guided abstractive summarisation architecture that uses MedCAT-extracted concept sequences to guide the text summariser. We use teacher-forcing for the MedCAT-extracted concept encoder input, and the decoder output embedding signal. Code is made available for the adapted BART model and the input preparation\footnote{https://github.com/tomolopolis/BHC-Summarisation/blob/master/guidance\_models/}.

We configure the guidance signals to include only \textit{problem} (disease, disorder, finding), and \textit{problem \& intervention (drug, procedure)} extracted concepts. This aims to explore the effect of varying the guidance signal across datasets. The original work indicates the guidance signal choice can affect the resulting summary performance\cite{Dou2021-lt}.

\subsection{Extractive and Abstractive Ensemble Model}

Our final experiments ensemble the above clinically guided abstractive model with our extractive top-level summary models, therefore utilising both the extractive and abstractive models simultaneously. We predict the initial \textit{n} lines of the BHC section using the extractive model then use the abstractive model with the guidance signal to predict the following sentences. Importantly, the ensemble predictions are fed into the abstractive model - to replicate the scenario of the summarisation model having already produced these sentences.

\section{Results}\label{sec:results}

Our results can be interpreted as the balance of generated summary relevancy, i.e. including only content found in the reference summary, and coverage, i.e. content available in the reference summary is in the generated summary. Prior work has shown a positive correlation between the higher the ROUGE scores the high performing summary when manually judged by a human\cite{Lin2004-my}.

\subsection{Extractive Models}
Our extractive models rank all sentences within the source text to find the top-k salient sentences that comprise the summary. Table \ref{tab:extractive_results} show our results across varying initial BHC section sentence limits for the various model embedding and ranking model configurations. Prior work found BHC sections are initially extractive then quickly move to abstractive problem focused narratives\cite{Adams2021-hp}. The Oracle model that has access to the target BHC section to rank candidate sentences against, shows the performance ceiling on both datasets is between 5 and 10 of the initial BHC sentences. This is more clearly shown in the KCH dataset with only a very small improvement between 5 and 15 sentences.

Our best performing ranker models use the semantic contextual sentence embeddings from S-BERT and the LSTM ranker across the majority of the sentence limits for both datasets. It is noteworthy that the improvements of using sentence specific embeddings S-BERT vs average word vectors are minor in comparison to performance improvements from the unsupervised TextRank ranker to the supervised LSTM model. This suggests that relying on relative importance of words and sentences within the documents is an ineffective model, and domain knowledge is needed to build BHCs.

\setlength\tabcolsep{3.5pt}
\begin{table*}[]
    \centering
    \begin{tabular}{c cccc|c|cccc|c}
    \toprule
    & \multicolumn{5}{c}{\textbf{MIMIC-III}} & \multicolumn{5}{c}{\textbf{KCH}}\\
    & \multicolumn{2}{c}{\textbf{TextRank}} & \multicolumn{2}{c}{\textbf{Bi-LSTM}} & & \multicolumn{2}{c}{\textbf{TextRank}} & \multicolumn{2}{c}{\textbf{Bi-LSTM}} &  \\
    \textbf{Sentence Limit} & \textbf{WV} & \textbf{SB} & \textbf{WV} & \textbf{SB} &\textbf{Oracle}& \textbf{WV} & \textbf{SB} & \textbf{WV} & \textbf{SB} & \textbf{Oracle} \\
    \midrule
    1         & 0.0         & 0.0        & 18.3        & \textbf{21.8}     & 30.2 & 4.09 & 3.6 & 4.3 & \textbf{14.7} & 23.3 \\
    2         & 5.6         & 5.0        & 17.2        & \textbf{18.8}     & 31.1 & 5.56 & 5.2 & 8.3 & \textbf{10.1} & 29.1\\
    3         & 7.6         & 5.1        & 16.6        & \textbf{17.5}     & 31.8 & 6.63 & 6.4 & \textbf{10.8} & 9.9 & 31.7   \\
    5         & 18.8         & 11.3        & 22.1        & \textbf{23.5}     & 32.8 & 7.61 & 7.5 & \textbf{16.1} & 12.4 & 34.2\\
    10        & 17.9         & 17.7        & 27.5        & \textbf{28.7}     & 34.3 & 9.12 & 9.2 & 13.4 & \textbf{20.59} & 35.6 \\
    15        & 24.1         & 28.3        & 30.1        & \textbf{31.1}     & 35.3 & 13.0 & 12.9 & 15.8 & \textbf{16.0} & 35.3 \\
    \bottomrule
    \end{tabular}
    \caption{ROUGE-LSum F1 scores for the extractive summarisation via sentence ranking for varying sentence limits. \textbf{WV} is the Word vector embedding method, and \textbf{SB} the sent-BERT embedding method used as input to our modelling approaches TextRank or Bi-LSTM. \textbf{Bold} indicates the best score across each sentence limit experiment. The Oracle model results are the performance ceiling for each configuration.}
    \label{tab:extractive_results}
\end{table*}

\subsection{Abstractive Models}
Table \ref{tab:pre-trained-abs-resutls} shows our pre-trained Transformer based models fine-tuned on our datasets. We observe that the performance of these deep pre-trained models is not comparable with open-domain summarisation, even when these models are further pre-trained on biomedical corpora such as PubMed or even MIMIC-III itself. Prior work reports ROUGE-L F1 scores between 37-41 points for BART, BERT, T5 with the open domain summarisation datasets, namely the CNN/Daily Mail\cite{Nallapati2016-ur} and XSum\cite{Narayan2018-hn} datasets, whereas our results show a range between 7-32 points. The ROUGE-2 performance gap is even larger with open-domain summarisation for these models varying between 19-22 and our results showing a range 1-11 points on our clinical datasets. The BART model pre-trained on PubMed is our best performing model by a substantial margin for both MIMIC-III and KCH BHC summarisation.

\begin{table}[]
    \centering
    \begin{tabular}{lcc}
        \toprule
        \textbf{Model} & \textbf{M-III} & \textbf{KCH}\\
        \midrule 
        T5-base       &     7.3  / 1.3 & 11.0 / 6.3 \\
        T5-small       &    14.4 / 5.6  & 10.8 / 4.1 \\
        BERT-2-BERT &       22.4 /  4.6  & 7.4 / 2.1  \\
        BERT-2-BERT (PubMed) & 23.8 / 4.2 & 6.2 / 1.6 \\
        BERT-2-BERT (M-III) & -     &  8.6 / 2.2 \\
        BART           &      26.9 / \textbf{11.1}  & 17.1 / 8.0 \\
        BART (PubMed)  &      \textbf{32.7} /  \textbf{11.1}  & \textbf{22.1} / \textbf{8.6} \\
        \bottomrule
    \end{tabular}
    \caption{ROUGE-LSum and ROUGE-2 F1 scores for pre-trained transformer models fine-tuned on the entirety of the BHC Summarisation task.}
    \label{tab:pre-trained-abs-resutls}
\end{table}

\subsection{Clinically Guided Abstractive Summarisation}
Table \ref{tab:guidance_aware_summarisers} shows our guidance aware abstractive summarisation results. We use 2 different guidance signals extracted by our pretrained MedCAT model. The first signal \textit{Prb} includes only the \textit{problem} extracted concepts. The second \textit{Prb + Inv} includes MedCAT extracted problems and interventions.

\setlength\tabcolsep{2pt}
\begin{table}[]
    \centering
    \begin{tabular}{lcc}
        \toprule
        \textbf{Model} & \textbf{M-III} & \textbf{KCH}\\
        \midrule
        BART            &      26.9 / 11.1   & 17.1 / 8.0 \\
        BART + Prb      &       26.0 / 9.1   &  23.4 / 12.0  \\
        BART + (Prb \& Inv) &  26.2 / 8.5   & 23.4 / 12.2  \\
        BART(PubMed)            &  32.7 / 11.1  & 22.1 / 8.6 \\
        BART(PubMed) + Prb      &   \textbf{34.7} / 10.6  &   \textbf{26.6} / \textbf{13.7} \\
        \shortstack{BART(PubMed) \\+ (Prb \& Inv)}  & 33.6 / \textbf{11.5} & 24.0 / 12.8    \\ 
        \bottomrule
    \end{tabular}
    \caption{ROUGE-LSum / ROUGE-2 F1 scores for our clinically guided abstractive summarisation models. BART is pre-trained on the open-domain XSUM\cite{Narayan2018-hn} datasets, and BART (PubMed) is pre-trained on PubMed\cite{Gupta2021-au}. \textbf{Bold} indicates the best performance for the metric and dataset}
    \label{tab:guidance_aware_summarisers}
\end{table}

The M-III BART shows a small drop in performance, 1 and 3 points with both guidance signals, whereas the KCH model improves by 6 and 4 points for ROUGE-LSum and ROUGE-2 respectively. 
For BART(PubMed) we observe improved ROUGE-LSum performance with both guidance signal types \textit{Prb} and \textit{Prb + Inv}. We observe a small gain with ROUGE-2 in MIMIC-III but more noticeable in KCH(~4 points). BART(PubMed) experiments show both guidance signals are comparable, with \textit{Prb} offering a marginal improvements when compared to the \textit{Prb + Inv} signal, despite there being less \textit{guidance} offered.

\subsection{Ensemble Extractive / Abstractive Summarisation}
Table \ref{tab:ensemble_summarisers} shows ablation results for our baseline and ensemble models. \textit{Abs} is the abstractive only model BART with PubMed pre-training. \textit{Ext + Abs} is the extractive and abstractive model - S-BERT into Bi-LSTM sentence ranker and BART with PubMed fine-tuning. \textit{Ext + Abs + Prb} is our final model that is extractive and abstractive with \textit{Problem} extracted clinical term guidance.

\setlength\tabcolsep{3pt}
\begin{table}[]
    \centering
    \begin{tabular}{lcc}
        \toprule
        \textbf{Model} & \textbf{M-III} & \textbf{KCH}\\
        \midrule 
        Abs           &    32.7 / \textbf{11.1} & 22.1 / \textbf{8.6} \\
        Ext + Abs  &     \textbf{34.9} / 10.6   & \textbf{23.6} / 7.5   \\
        Ext + Abs + Prb &  \textbf{34.9} / 10.6  & 22.4 / 6.7       \\
        \bottomrule
    \end{tabular}
    \caption{ROUGE-LSum and ROUGE-2 F1 score results for our baseline abstractive and ensemble summariser configurations. \textit{Bold} indicates the best performance for the respective metric / dataset pair.}
    \label{tab:ensemble_summarisers}
\end{table}

We only observe small improved performance through either ensembling with or without guidance. Only the KCH ROUGE-2 score is worse with the ensemble model. 

\subsection{Summarisation Extracted Concept Analysis}
Alongside ROUGE scores, we analyse the clinical terms output by our summarisation models. As our guidance signal should push the model to generate more clinically relevant information. We run our pre-trained NER+L model (MedCAT), the same model used to produce the guidance signals, over the generated summaries from the models in Table \ref{tab:ensemble_summarisers} comparing the proportion of terms in the generated vs reference summary.

Appendix Table \ref{appdx_tab:mc_extracted_sum_concepts} provides full results. There are small improvement with both datasets using the guidance model, with summaries having 0-4\% more clinical terms in the guidance models compared to the baseline abstractive models, indicating the guidance signal is assisting the model produce more clinically relevant terms. The guidance assists the generation of problems more so than interventions unsurprisingly as this guidance only includes problem extracted terms. Overall, there is still a majority of concepts (>50\%) that are missed entirely by all generated summaries, suggesting there is plenty of room for improvement.

\subsection{Qualitative Analysis}
We manually review 40 random summaries from the set of model configurations presented in Table \ref{tab:ensemble_summarisers} with two clinicians. We compare the generated BHC, the reference summary and the original source notes for only the MIMIC-III dataset due to the sensitivity of the KCH data. Examples of these comparisons can be found in Appendix C. We use a Likert scale 1-5, to measure: 1) coherence - the overall quality of all sentences of the BHC, 2) fluency - the quality of each individual sentence, 3) consistency - the correct facts are in the BHC compared to source notes, 4) relevance - the BHC only contains the relevant facts from the source notes and is not overly verbose. These measures have been used and defined in prior work for large scale qualitative assessment of summarisation texts\cite{Fabbri2021-ka}. Our reviewers - review BHCs blind as to not bias the ratings towards either the reference or generated summaries. We record an average Cohen's Kappa of 0.65 across the 4 metrics. We take the mid point score if there are disagreements. From all ratings there are no disagreements larger than 2 points.

Now we discuss the \% of samples with scores $\geq 2.5$ for each metric. For \textit{coherence} we observe that all our models achieve 70\% (28/40) vs 75\% (30/40) for the reference summary. For \textit{fluency} our models achieve 60\% (24/40) vs 70\%(28/40) for reference summaries. For \textit{consistency} we see a small difference in favour of our guidance model 58\% (23/40) vs abstractive baseline 55\% (27/40). Reference summary consistency was at 90\% (36/40). Finally, \textit{relevance} showed another small improvement with the guidance models 73\%(29/40) vs 70\%(28/40) with the reference summary at 80\%(32/40).

We find that the majority of the same summaries are rated $\geq 2.5$ over the 4 metrics. Indicating the variability in difficulty the models encountered with the task. Overall, from this small scale manual analysis it is positive to see that the models, including the baseline abstractive model, performing well across all metrics. However, there is still much room for improvements, with between 30\% - 40\% of produced summaries without acceptable outputs. This poor performing text resulted in common abstractive summarisation issues such as repeated phrases or words within and across sentences, and most worryingly are the occurrences of inconsistencies between source and generated summary facts. For example, `No documented hypoxia at this hospital' is correctly within the reference but is generated in the BHC summaries: `Hypoxia:  The patient was initially hypoxic on admission to the ICU.'. This inconsistent fact is between multiple consistent facts in the generated summaries. A high performing summary must be near perfect in its consistency to be usable in a real scenario.

A promising result here is that these bad performing summaries were often easy to pick out, and could potentially be systematically excluded if the model were to be included within a real production system. For example the system could decline to `auto-complete' a summary given a set of admission notes, if the produced summary excessively repeated a phrase or sentence.

We notice that our ensembling strategy to first sample extractive sentences then from the abstractive model do not read as coherently as the abstractive only models. This indicates that summaries move between extractive and abstractive generation at the sub-sentence level, and require a more sophisticated model to balance extractive selection of representative words or phrases alongside abstractive generation, e.g. the Pointer Generator model\cite{See2017-fx}.

\section{Discussion and Future Work} 
We first discuss our initial baseline methods - our extractive models and our pre-trained fine-tuned abstractive models. We then discuss our guidance signal enhanced model and final ensemble approach.  We then discuss a range of issues of our approaches and the problem more broadly. This includes common problems with abstractive models, reference summary quality and the difficulties around real-world clinical text, summarisation metrics and possible future directions for real-world usage of such systems.

Our sentence ranking extractive summarisation experiments suggest the amount of `extractiveness' for a BHC section depends largely on the dataset. Prior work showed that BHC sections often rely on extractive summarisation initially, i.e. direct copy and paste from source notes into the BHC for the first few sentences, but then quickly switch to abstractive summarisation in later sentences\cite{Adams2021-hp}. Our work supports the finding that both extractive and abstractive techniques are used. The M-III dataset shows the opening sentences of the BHCs are more consistently `extractive' than KCH, as seen by the differences in Oracle model performance as the sentence limit increases. Our best performing model uses a pre-trained contextual sentence embedding model (S-BERT) alongside a Bi-LSTM. Future work could consider further ranking models i.e. a Transformer model to rank sentences, or an appropriate embedding boundary to build sub-sentence, or phrase level embeddings extractive summaries from these. Moreover, we would expect to see different results in sub-sentence level extractions over the whole sentence extractions that we report.

Our fine-tuning of pre-trained abstractive summarisation systems suggest BART, the only model specifically trained for NLG tasks such as summarisation, offers the best performance across datasets and metrics for BHC summarisation. Models such as T5, a general seq-to-seq Transformer model and the BERT-2-BERT models perform substantially worse than BART. For BART we find that further pre-training on a relevant corpus i.e. PubMed\cite{Gupta2021-au} compared to only open-domain pre-training, offers improvements inline with prior research\cite{Rogers2020-nl}.

We find that guidance signals for BHC abstractive summarisation offers improvements compared to our best model without guidance. We observe best performance once an existing pre-trained model has already been fine-tuned with biomedical data. We observe that guidance signal improvements are dataset dependent. All experiments use the equivalent hyperparameters, e.g. learning rate, learning rate scheduler, epoch number etc. as the baseline abstractive models. It is likely that further performance gains are possible with further hyperparameter tuning. The guidance models share the parameters for the initial 3 encoder layers. Further work could explore the effect of increasing or decreasing the number of shared parameters.

\subsection{Guidance Signal}
The guidance signal uses a pre-trained MedCAT\cite{Kraljevic2021-ln} model. This model has not been validated across the entirety of clinical terms that could be extracted. It has been configured to favour precision over recall, and so likely misses clinical terms that otherwise should be identified and included within the guidance signal. Further work could fine-tune and improve the model performance to improve the guidance signal offered to the summarisation model using the MedCAT annotation tool and workflow\cite{Searle2019-fd}. 

This guidance signal used in our experiments is produced using the MedCAT\cite{Kraljevic2021-ln} NER+L approach that is trained unsupervised on the same MIMIC-III and KCH text data. This approach could be replaced with a rules-based, ML or otherwise approach to extract clinical terms. The effectiveness of the clinical term extraction and subsequent usage as a guidance signal will impact the effectiveness of the adapted model. If the NER+L sufficiently under performs and relevant terms are missed, it is very likely the guidance assisted model will perform the same or worse than the standard abstractive model as the decoder stack needs to learn to ignore cross-attention from the encoder.

Moreover, it must be highlighted that our NER+L model has seen the MIMIC-III / KCH data during unsupervised training although it has has not received supervised training on any of this data. MedCAT models are based upon a concept dictionary lookup, alongside a concept vector disambiguation algorithm that adapts concept vectors according to the context in which they are found. We have configured the model to highly favour high confidence predictions (i.e. high precision) predictions so  it is likely that the majority of predictions are simply dictionary matches. However, the impact of pre-training this guidance model has on our results is unclear and should be considered alongside our results. The guidance signal could be biased and higher performing than signal output by a model that has not seen the input summarisation data. Overall, As reported in prior work\cite{Dou2021-lt}, further work on guidance signal generation is needed.

For successful model fine-tuning the guidance signal must be aligned with the raw text input. We align the signal by padding the signal with the white space token, but further experiments could investigate aligning the signal with syntactic hints such as punctuation, i.e. full stops, commas, new lines, colons etc. Further work could also experiment with replacing identified guidance terms directly with clinical concept embeddings. During our experiments we attempted to replace the raw text with the standardised terminology name but this lead to model failures and only keeping the original source text allowed for model convergence.

\subsection{Ensemble Models}
We use a very simple ensembling strategy, sampling the extractive model and feeding into the abstractive summariser. Prior work suggests that BHC sections are initially extractive then become abstractive\cite{Adams2021-hp}. We find this to be partly true - we reach an Oracle performance limit for both datasets between 10 and 15 sentences - but it is probably at the sub-sentence / phrase level rather than full sentences where summaries are extractive. Further work could explore a PG\cite{See2017-fx} network architecture, with a mechanism to favour extractiveness initially then abstractive generation afterwards. 
 
\subsection{Problems with Abstractive Summarisation Models}
Repetition is a known problem with Abstractive summarisation models\cite{Nair2021-zf}. Prior work have studied numerous methods to reduce repetition and therefore improve summarisation quality.   These include a specific training regime that improves the models ability to sample previously unselected n-grams\cite{Welleck2020-uc}, and a coverage model that adjusts the loss to include words and phrases that sufficiently cover the source text\cite{See2017-fx}. Repetition is highly unlikely to occur in human generated summaries. Utilising the above techniques would likely improve performance, as observed in open-domain settings\cite{Nair2021-zf}, although we argue this would still not guide the model to `focus' on the problem-list during summary generation as our method allows.

Factual correctness is an important problem in summarisation and especially important applying these models to clinical scenarios, a high stakes use case that lead to large downstream impacts for model errors. An incorrect statement within a generated BHC summary could miss a diagnosis, follow-up or report a result incorrectly. Our own manual analysis identified various examples especially within long BHCs, of occurrences of inconsistent facts, detecting these and ensuring the model is consistent with the source text is arguably the most important metric in BHC generation. A real deployment of a BHC summarisation system would likely require a `human-in-the-loop' to monitor, similar to most medical AI\cite{Jotterand2020-vc}. The human user would correct, further edit and sign-off on any produced summaries. Even if a system were only able to provide a basic BHC summary, this would still beneficially reduce the administrative burden of completing the BHC section from scratch.

\subsection{Reference Summary Quality}
The reference summary BHC sections in both datasets were collected as part of routine care. They have not been reviewed and validated so represent a silver-standard BHCs. Real-world clinical data often does not undergo secondary validation, and even MIMIC-III a heavily studied clinical dataset has data quality concerns\cite{Searle2020-xb,Afshar2021-lh}. It is likely that there are mistakes and omissions in this dataset but given the complexity of clinical text, developing a gold-standard double annotated corpora would be prohibitively expensive. If for example we used two clinicians and each took on average 30 minutes per admission, one to generate a new summary of each admission, and another to compare both the existing reference and newly written summary this would still take circa. 7.4 years of manual work for both clinicians, of 8 hour work days, 5 days a week and 40 weeks a year. This is clearly not going to be possible, across multiple datasets.

However, we argue in line with prior analysis that BHC writing is context and author specific so it is likely another domain expert clinician with different training, geography etc. would result in a different summary\cite{Adams2021-hp}. Future work could seek to better understand the variability between BHC sections, or even validate BHC sections creating a gold-standard. 

\subsection{Summarisation Metrics}
The ROUGE score shows our guidance assisted and ensemble models offer some but limited improvement. However, in context current top performing ROUGE-LSum scores in open-domain summarisation are at 37-41 points\cite{Lewis2020-pw} and improvements needed for achieving a few points above the current state-of-the-art is difficult. Analysis using MedCAT extracted concepts and from manual review indicates the addition of guidance helps to produce longer more `clinically complete' summaries despite the similarity in ROUGE score.

ROUGE has been criticised in the literature as summarisation quality can score highly whilst perform poorly during manual evaluation\cite{Schluter2017-ov}. Alternative metrics such as BERTScore\cite{Zhang2020-lf} or the recently introduced question answering metrics\cite{Eyal2019-zl,Wang2020-ay} rely on manually generating questions for reference/generated summary pairs or a pre-trained answer conditional question generation model. Assessing our experimental scenarios with these metrics is left to future work, but would likely assist in higher quality, more factually correct summaries. Factual accuracy is critical in BHC generation, as this section is both a legal record and likely to be used by followup care upon discharge.

\subsection{Downstream Summary Use}
Automatic generation of BHC sections from source notes is still a long way off. Embedding an automatic summarisation model in \textit{high stakes scenarios} such as healthcare would involve engineering a solution well beyond a research project. Aside from initial validation, ML operations tasks such as detecting model drift or bias would be essential. 

In any real-use scenario - a generated summary would likely only be used with explicit clinician supervision and ultimate responsibility for the produced summary, ensuring factual correctness and coherence. \cite{Pivovarov2015-nw} provides a further categorization of generated summaries and how the output is integrated into a workflow. They explain that \textit{indicative} summaries highlight significant or important parts of source texts, whereas \textit{informative} summaries are intended to replace the original text and used in place of it.

\section{Conclusions}
Our work has demonstrated a range of possible models using both extractive, abstractive summarisation approaches, pre-trained and fine-tuned to specific data and a pre-trained guidance signal generation model (MedCAT) to push the summarisation models to focus on clinically relevant terms. We train a state-of-the-art abstractive model guided by clinically relevant \textit{problem} terms outperforming all baselines across 2 real-world clinical dataset.

Overall, we have shown BHC automated summarisation to be a challenging task supporting prior work\cite{Adams2021-hp} suggesting that BHCs are both extractive and abstractive. We hope this work motivates further work in this area that could one day improve the overall healthcare experience for patient and clinician alike through the minimisation of \textit{screen time}. A well documented contributing factor for clinician burn-out\cite{McPeek-Hinz2021-nz,ODonnell2009-ye}. 

\section*{Acknowledgements}
RD's work is supported by 1.National Institute for Health Research (NIHR) Biomedical Research Centre at South London and Maudsley NHS Foundation Trust and King’s College London. 2. Health Data Research UK, which is funded by the UK Medical Research Council, Engineering and Physical Sciences Research Council, Economic and Social Research Council, Department of Health and Social Care (England), Chief Scientist Office of the Scottish Government Health and Social Care Directorates, Health and Social Care Research and Development Division (Welsh Government), Public Health Agency (Northern Ireland), British Heart Foundation and Wellcome Trust. 3. The National Institute for Health Research University College London Hospitals Biomedical Research Centre. This paper represents independent research part funded by the National Institute for Health Research (NIHR) Biomedical Research Centre at South London and Maudsley NHS Foundation Trust and King’s College London. The views expressed are those of the author(s) and not necessarily those of the NHS, MRC, NIHR or the Department of Health and Social Care.




\bibliography{main}
\bibliographystyle{main}

\appendix

\section{MedCAT Extracted Terms}
We configure MedCAT\cite{Kraljevic2021-ln} to extract `problem' and intervention terms. Table \ref{tab:prob_terms} and \ref{tab:interven_terms} are provided.

\begin{table*}[]
    \centering
    \begin{tabular}{cccc}
    \toprule
    \textbf{Type ID} & \textbf{SCTID Root Term} & \textbf{Description} & \textbf{\# Concepts Available}\\ \midrule
    T-11    & 64572001               & Disorder & 77,284 \\
    T-18    & 404684003              & Clinical Finding  & 44,201 \\
    T-29    & 49755003               & Morphologic Abnormality & 4,897 \\
    T-35    & 410607006              & Organism & 34,778 \\
    T-38    & 260787004              & Physical Object & 198,890 \\ \bottomrule
    \end{tabular}
    \caption{The set of `Problem' semantic tags from SNOMED-CT configured within MedCAT, and extracted from source texts and BHC summaries}
    \label{tab:prob_terms}
\end{table*}

\begin{table*}[]
    \centering
    \begin{tabular}{cccc}
    \toprule
    \textbf{Type ID} & \textbf{SCTID Root Term} & \textbf{Description} & \textbf{\# Concepts Available}\\ \midrule
    T-9     & 373873005       & Clinical Drug          & 6,247 \\
    T-26    & 373873005       & Medicinal Product      & 7,715 \\
    T-27    & 373873005       & Medicinal Product Form & 6,203 \\
    T-39    & 71388002        & Procedure              & 6,4291 \\
    T-40    & 373873005       & Product                & 17,3894 \\
    T-55    & 105590001       & Substance              & 27,626 \\ \bottomrule
    \end{tabular}
    \caption{The set of `Intervention' semantic tags from SNOMED-CT configured within MedCAT. All SNOMED-CT terms with these semantic terms are extracted from source texts and BHC summaries and treated as `Intervention' terms.}
    \label{tab:interven_terms}
\end{table*}

Table \ref{tab:medcat_desc_stats} shows descriptive statistics of the extracted terms of both datasets MIMIC-III and KCH. 

\begin{table}[]
    \centering
    \begin{tabular}{clcc}
    \toprule
                                    &                    & \multicolumn{2}{c}{\textbf{Dataset}} \\
                                    &                    & \textbf{M-III}     & \textbf{KCH}    \\\midrule
    \multirow{4}{*}{\textbf{Notes}} & \# Terms           & 156                &     110            \\
                                    & Term Density       & 55                 &     26           \\
                                    & \# Uniq Terms      & 56                 &     43           \\
                                    & Uniq Term Density & 118                &      32          \\
                                    \midrule
    \multirow{4}{*}{\textbf{BHC}}   & \# Terms           & 19                 &     10            \\
                                    & Term Density      & 52                  &     29           \\
                                    & \# Uniq Terms      & 15                 &     8           \\
                                    & Uniq Term Density  & 63                 &     35  \\\bottomrule
    \end{tabular}
    \caption{Extracted and linked average: term counts, unique term counts, and their respective densities with regards to the number tokens per clinical term.}
    \label{tab:medcat_desc_stats}
\end{table}

\section{Extractive Baseline Architectures}
\begin{figure}
    \centering
    \includegraphics[scale=0.2]{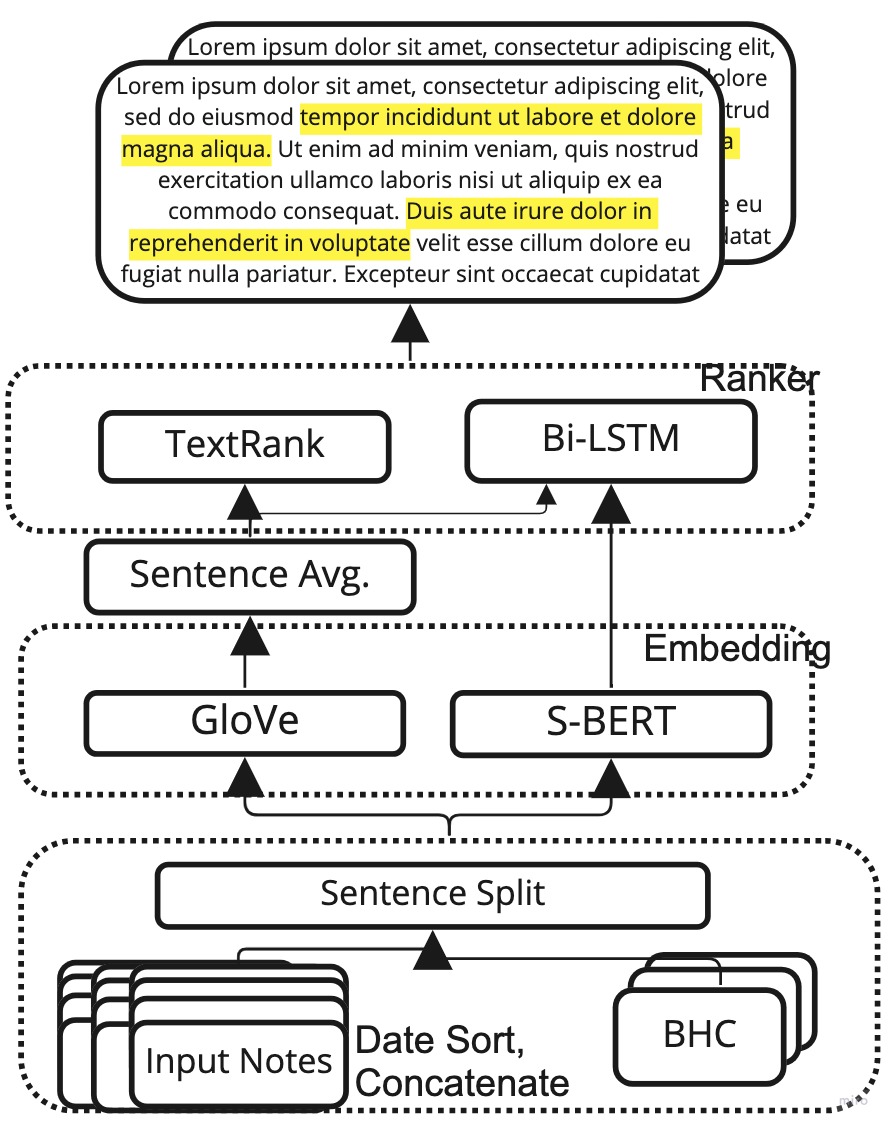} 
    \caption{Baseline Top-N sentence extractive model architectures}
    \label{fig:extractive_models}
\end{figure}

Fig. \ref{fig:extractive_models}, shows our baseline extractive model architectures. 

\section{Extractive Summarisation Plots Precision, Recall, F1 Plots Appendix}
Extractive Summarisation Methods for Top-N-Line BHC Summarisation 
\label{sec:}

\begin{figure*}
    \centering
    \includegraphics[scale=.5]{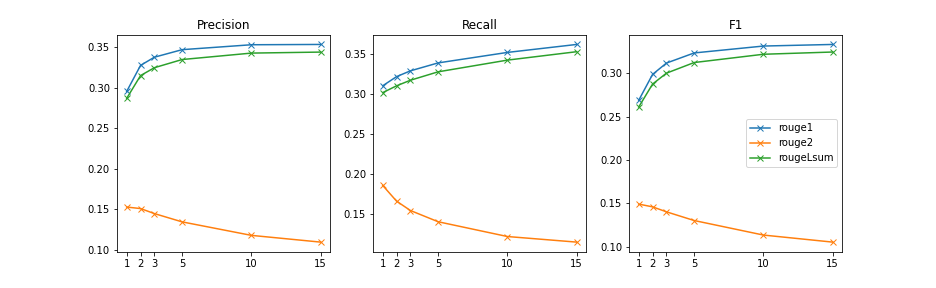}
    \caption{Extractive score max}
    \label{fig:oracle}
\end{figure*}

\begin{figure*}
    \centering
    \includegraphics[scale=.5]{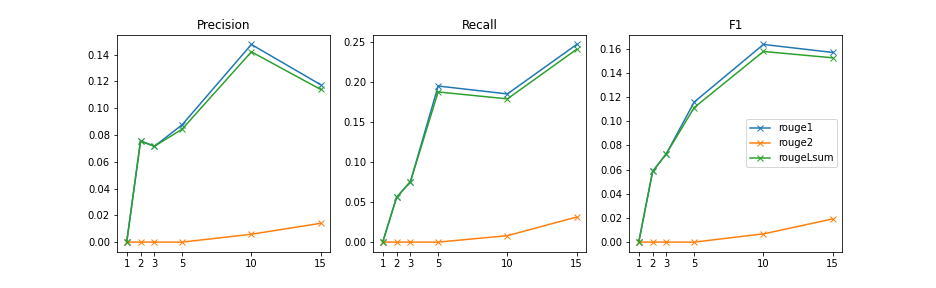}
    \caption{GloVe Embeddings: TextRank}
    \label{fig:wv-tr}
\end{figure*}

\begin{figure*}
    \centering
    \includegraphics[scale=.5]{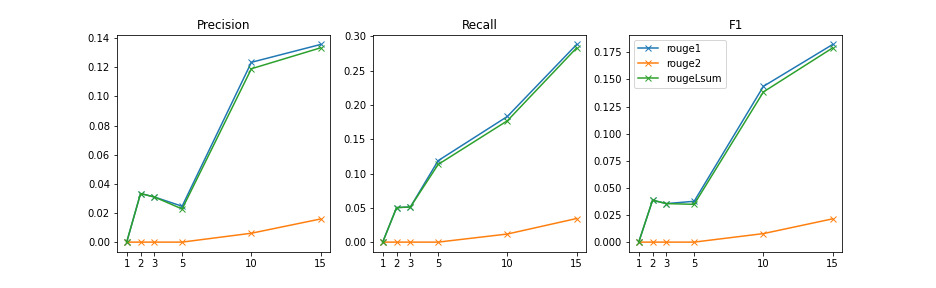}
    \caption{S-BERT embeddings: TextRank }
    \label{fig:sbert-tr}
\end{figure*}

\begin{figure*}
    \centering
    \includegraphics[scale=.5]{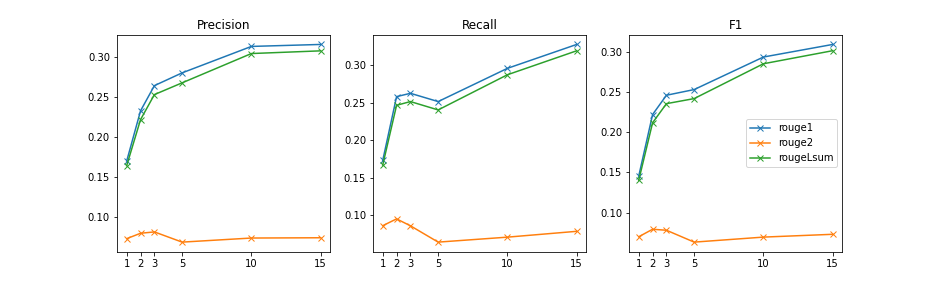}
    \caption{S-BERT embeddings: Bi-LSTM}
    \label{fig:sbert-bilstm}
\end{figure*}

\section{Measures of Clinically Relevant Information Across Summarisation Models}
Table \ref{appdx_tab:mc_extracted_sum_concepts} shows the proportion of concepts that we successfully generate in the predicted summaries vs the reference summaries. 

\setlength\tabcolsep{1.5pt}
\begin{table}[]
    \begin{tabular}{@{}clccc@{}}
    \toprule
        \textbf{Dataset}  & \textbf{Model}   & \textbf{\% Prob} & \textbf{\% Inv} & \textbf{\% Total}\\ \midrule
    \multirow{4}{*}{M-III} & Abs             &  31    & 32       & 34\\
                           & Ext + Abs       &  33    & 33       & 34 \\
                           & Ext + Abs + Prb   &  \textbf{35}    & \textbf{35}       & 34 \\ \midrule
    \multirow{4}{*}{KCH}    & Abs             &  40   & 30     & 38 \\
                           & Ext + Abs       &  41   & \textbf{34}     & 41 \\
                           & Ext + Abs + Prb  &  \textbf{43}   & \textbf{34}     & \textbf{42} \\\bottomrule 
    \end{tabular}
    \caption{MedCAT Extracted Term comparisons vs reference summary. Average \% of problem only, intervention only and both problem \& intervention terms in the generated vs the reference summary. \textbf{Bold} indicates model with highest proportion of clinical terms generated compared with reference summary.}
    \label{appdx_tab:mc_extracted_sum_concepts}
\end{table}

\end{document}